\documentclass[letterpaper, 10 pt, conference]{ieeeconf}  % Comment this line out if you need a4paper
\usepackage{xcolor}
\usepackage{bm}
\usepackage{times}
\usepackage{graphicx}
\usepackage{amsmath}
\usepackage{amssymb}
\usepackage{gensymb}
\usepackage{pifont}% http://ctan.org/pkg/pifont
\usepackage{multicol}
\usepackage{hhline}
\usepackage{acro}
\DeclareRobustCommand{\rchi}{{\mathpalette\irchi\relax}}
\newcommand{\irchi}[2]{\raisebox{\depth}{$#1\chi$}}

\usepackage[sorting=none]{biblatex}
\bibliography{mybibliography}

\IEEEoverridecommandlockouts                              % This command is only needed if 
\newcommand{\cmark}{\ding{51}}%
\newcommand{\xmark}{\ding{55}}%
\usepackage{graphics}

\title{\LARGE \bf
BEV-CV: Birds-Eye-View Transform for Cross-View Geo-Localisation
}

\author{Tavis Shore$^{1}$ and Simon Hadfield$^{1}$ and Oscar Mendez$^{1}$% <-this % stops a space
\thanks{$^{1}$Tavis Shore, Dr Simon Hadfield, and Dr Oscar Mendez are with the Centre for Vision, Speech, and Signal Processing (CVSSP),
        University of Surrey, Guildford, United Kingdom
        {\tt\small {initial}.{surname}@surrey.ac.uk}}}

\DeclareAcronym{gnss}{
    short = GNSS,
    long  = Global Navigation Satellite Systems,
    tag   = nomencl
}
\DeclareAcronym{cvgl}{
    short = CVGL,
    long  = Cross-View Geo-localisation,
    tag   = nomencl
}
\DeclareAcronym{flops}{
    short = FLOPs,
    long  = floating point operations,
    tag   = nomencl
}

\begin{document}

\maketitle
\thispagestyle{empty}
\pagestyle{empty}

%%%%%%%%%%%%%%%%%%%%%%%%%%%%%%%%%%%%%%%%%%%%%%%%%%%%%%%%%%%%%%%%%%%%%%%%%%%%%%%%
\begin{abstract}

Cross-view image matching for geo-localisation is a challenging problem due to the significant visual difference between aerial and ground-level viewpoints. The method provides localisation capabilities from geo-referenced images, eliminating the need for external devices or costly equipment. This enhances the capacity of agents to autonomously determine their position, navigate, and operate effectively in GNSS-denied environments.
Current research employs a variety of techniques to reduce the domain gap such as applying polar transforms to aerial images or synthesising between perspectives. However, these approaches generally rely on having a $\bm{360\degree}$ field of view, limiting real-world feasibility.
We propose BEV-CV, an approach introducing two key novelties with a focus on improving the real-world viability of cross-view geo-localisation. Firstly bringing ground-level images into a semantic Birds-Eye-View before matching embeddings, allowing for direct comparison with aerial image representations. Secondly, we adapt datasets into application realistic format - limited Field-of-View images aligned to vehicle direction.
BEV-CV achieves state-of-the-art recall accuracies, improving Top-1 rates of $\bm{70\degree}$ crops of CVUSA and CVACT by 23\% and 24\% respectively.
Also decreasing computational requirements by reducing floating point operations to below previous works, and decreasing embedding dimensionality by 33\% - together allowing for faster localisation capabilities.

\end{abstract}

%%%%%%%%%%%%%%%%%%%%%%%%%%%%%%%%%%%%%%%%%%%%%%%%%%%%%%%%%%%%%%%%%%%%%%%%%%%%%%%%
\section{INTRODUCTION}

Localisation is necessary for many robotics applications - from autonomous vehicles to driverless railways the ability to localise must be ingrained. The majority of current localisation techniques rely on external sensors supplying either positional context or a calculated location. This reliance on external devices such as \ac{gnss} may lead to issues caused by occlusions or sensor errors which inhibit localisation. Similarly, LIDAR-based approaches are expensive and both power and data-hungry. Vision-based localisation offers a solution as cameras are low-cost and compact, enabling robotics to discover more information about their environment from which to self-localise. Moreover, most modern vehicles are equipped with forward-facing cameras, making the adoption of limited Field-of-View (FOV) \ac{cvgl} straightforward.

\begin{figure}[t!]
    \centering
    \includegraphics[width=\columnwidth]{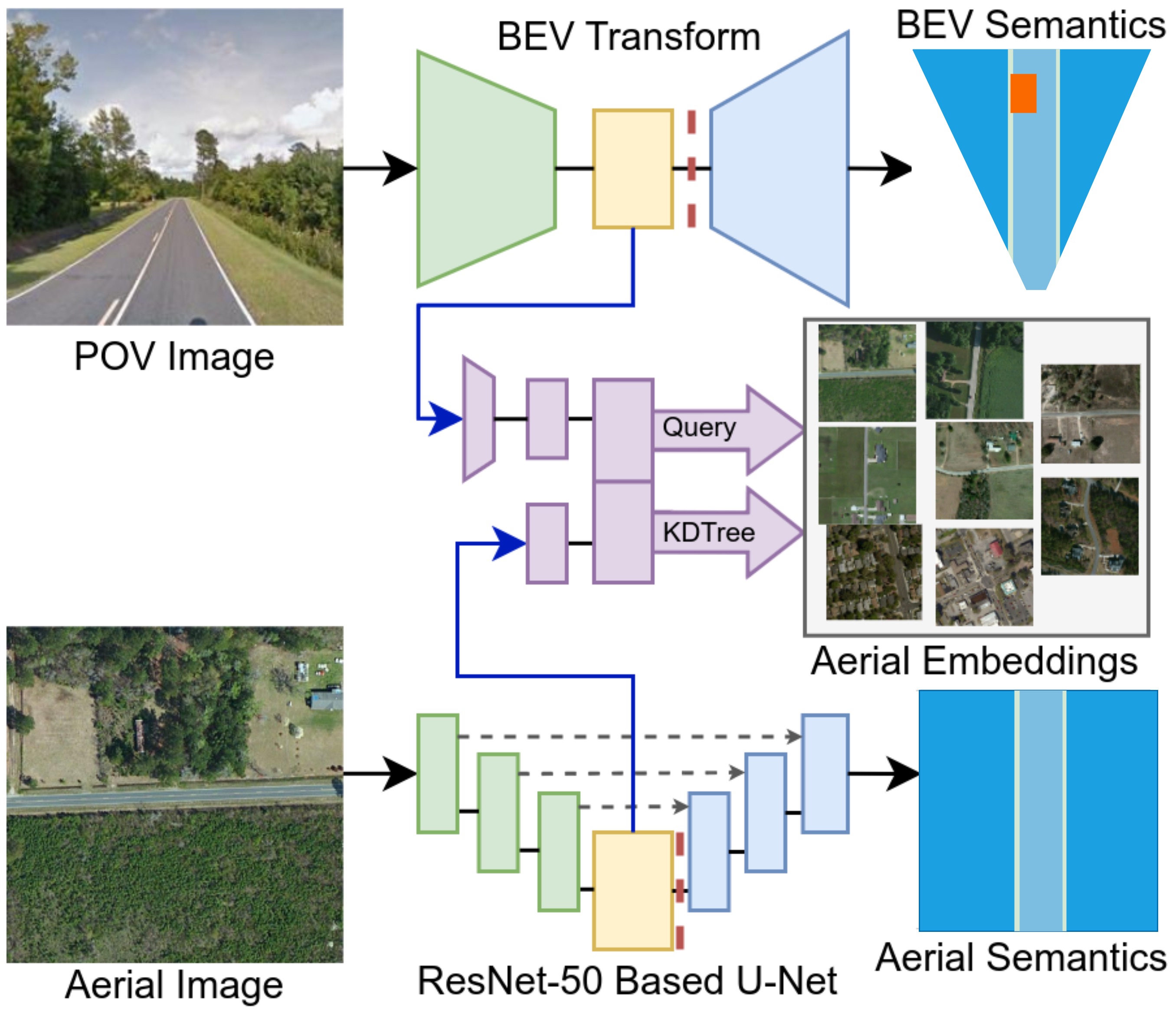}
    \caption{General BEV-CV network structure. \textit{POV Branch} extracts and transforms ground-level feature embeddings, \textit{Map Branch} extracts aerial embeddings to build a KDTree. Components to the right of dotted red lines are discarded in the final BEV-CV architecture.}
    \label{fig:overview}
    \vspace{-0.3em}
\end{figure}

% Considering images from different perspectives can add detailed location information. 
\ac{cvgl} aims to match ground-level perspective images to geo-referenced aerial images. Throughout this research, we refer to images taken from car-mounted front-facing limited-FOV cameras as Point-of-View (POV) images, and satellite or aerial images as aerial images.
\ac{cvgl} may offer a solution for self-contained localisation as a database of aerial feature embeddings is created locally and continually queried with POV feature embeddings, shown in Figure \ref{fig:overview}. 
% Localisation accuracy is only restricted by the system's ability to successfully match images.

Our research develops a novel approach to reduce the domain disparity between POV and aerial images, improving \ac{cvgl} performance while reducing computational requirements. We introduce BEV-CV, an architecture that reduces the domain gap between POV and aerial images by extracting semantic features at multiple resolutions before projecting them into a shared representation space and matching embedding pairs. We prioritise limiting computational requirements to ensure viability for mobile robotics applications. 

\vspace{0.1em}
In summary, technical contributions of our research are:
\begin{itemize}
    \item Novel multi-branch architecture for extracting top-down representations from both viewpoints, projecting these into a shared representation space. 
    \item Improving \ac{cvgl} feasibility in two aspects: adjusting benchmark datasets to closer represent real-world application, and focusing on developing an efficient system capable of running in mobile systems - approximately reducing query times by 18\% and embedding database memory requirements by 33\%.
    \item  Evaluation of our approach on CVUSA and CVACT datasets shows relative improvements upon previous state-of-the-art (SOTA) recall Top-1 rates by approximately 23\% for CVUSA and 24\% for CVACT with road-aligned 70$\degree$ crops.
\end{itemize}

\section{Related Works}
Most existing works focus on retrieving embeddings from aerial feature databases. The field has recently started considering limited-FOV images due to their wider applicability. We describe previous works which led to the development of BEV-CV.

\subsection{Cross-View Image Geo-Localisation}
%## Introduction
The predominant technique for CVGL is embedding retrieval. At an increasing rate, techniques are being proposed to reduce the domain gap and better match across viewpoints \cite{shi2020beyond}, \cite{Shi2020WhereAI}, \cite{cvlnet}. CVGL's deep learning era began in 2015 - Workman and Jacobs \cite{7301385} proposed CNNs for extracting relevant features from both viewpoints for comparison, achieving good results and proving the potential of neural networks in CVGL. CNNs have since remained the predominant feature extraction mechanism. Lin et al. \cite{7299135} considered each query image to have a unique identifier, using the cosine similarity distance as a similarity metric on clustered feature embeddings. Workman et al. \cite{7410808} extend this, embedding representations from both viewpoints into a joint semantic representation space. Vo and Hays \cite{Vo2016LocalizingAO} incorporated aerial rotational information through an auxiliary loss function to observe the impact of image pair misalignment, also introducing a distance-based logistic loss to optimise performance. Limited-FOV images worsen the impact from misalignment as there is less shared information between views with which to correlate. CVM-Net \cite{8578856} made a significant advancement, following a Siamese CNN with NetVLAD \cite{netvlad} - a network which aggregates residuals of local features to cluster centroids. \\
Toker et al. \cite{Toker2021ComingDT} use a multitask network that both synthesises streetview images from aerial image queries and performs image retrieval. Zhu et al. \cite{Zhu2020RevisitingSV} employ more recent metric learning techniques and leverage activation maps to perform orientation estimation. L2LTR \cite{Yang2021CrossviewGW} proposes a CNN+transformer architecture, a ResNet feature extractor with vanilla ViT encoder. TransGeo \cite{Zhu2022TransGeoTI} propose a transformer architecture that doesn't require data augmentation or alteration, using an attention-guided non-uniform cropping strategy to remove uninformative areas. \\
Sun et al. \cite{Sun2019GEOCAPSNETGT} propose a capsule network following a ResNet-based feature extractor, outperforming CVM-Net with the Vo and Hays dataset by approximately 10\%.  Liu and Li \cite{8954224} improved the representation ability of their latent space by inserting orientation information. Shi et al. \cite{Shi2019SpatialAwareFA} apply a polar transform to aerial images and develop a spatial attention mechanism to improve feature alignment between views. Regmi et al. \cite{Regmi2019BridgingTD} created a conditional GAN to synthesise aerial representations of ground-level panoramas to reduce the domain gap.
Shi et al. \cite{Shi2019OptimalFT} propose CVFT to perform cross-view domain transfer to improve feature alignment between images. Their subsequent paper \cite{Shi2020WhereAI} proposed the new problem of applying CVGL techniques to limited FOV crops. This is important due to the ubiquity of monocular cameras compared to panoramic cameras; the ability for CVGL to perform with a limited-FOV is essential for wide-spread feasibility and adoption. To resolve the orientation ambiguity \cite{Shi2020WhereAI} computes the feature correlation between ground-level images and the corresponding polar-transformed aerial image, shifting or cropping the panorama at the strongest point before matching with their Dynamic Similarity Matching (DSM) module. However, it does so at the cost of introducing an expensive post-processing step at both training and inference time. In contrast, the BEV derived features in our own approach are discriminative enough to match uniquely over orientations with no correlation-based post-processing. In GeoDTR \cite{zhang2023crossview} Zhang et al. disentangle geometric information from raw features, learning spatial correlations among visual features to increase performance.

\subsection{BEV Estimation}
BEV transforms are a prominent research sub-field, mainly for autonomous road vehicles. We summarise relevant BEV papers for theoretical background as our proposed technique is the first to apply the transform to large-scale CVGL. Traditional techniques used camera intrinsics to geometrically transform POV images into the horizontal plane, but these performed poorly in some object detection cases due to extreme warping. Lu et al. \cite{Lu2018MonocularSO} propose a variational encoder-decoder to encode ground-level images, decoding them into a semantic occupancy grid map. Schulter et al. \cite{Schulter2018LearningTL} introduce a CNN to predict occluded areas of a scene layout, negating labelling requirements for these portions. Roddick and Cipolla \cite{Roddick2020PredictingSM} propose a BEV network that extracts image features at multiple resolutions before augmenting with spatial context and mapping to the BEV space. Saha et al. \cite{Saha2021EnablingSA} use factorised 3D convolutions to estimate the BEV occupancy grid. Yang et al.  \cite{Yang2021ProjectingYV} employ a cross-view transformation module, using correlations to cyclically strengthen the view transformation. Saha et al. \cite{Saha2021TranslatingII} exploit the relationship between a vertical line in a POV image and a polar ray from the camera's perspective from the BEV viewpoint. Their subsequent research \cite{Saha2022ThePN} proposes a graph neural network to learn the spatial relationship between objects in a scene and improve BEV object estimation.

To our knowledge, only one piece of research applies a BEV transform to the cross-image field - Fervers et al. \cite{uncertainty} build a BEV representation of POV images with a cross-attention mechanism to determine a vehicle's pose. They use cross-attention to build output probabilities across a sequence of images, they could not evaluate with standard CVGL benchmark datasets as their technique requires lidar point clouds or rigid transformations between frames. BEV-CV is more applicable to wide-scale CVGL as point clouds are not required, only a single RGB perspective image.

\begin{figure*}
    \centering
    \includegraphics[width=\textwidth]{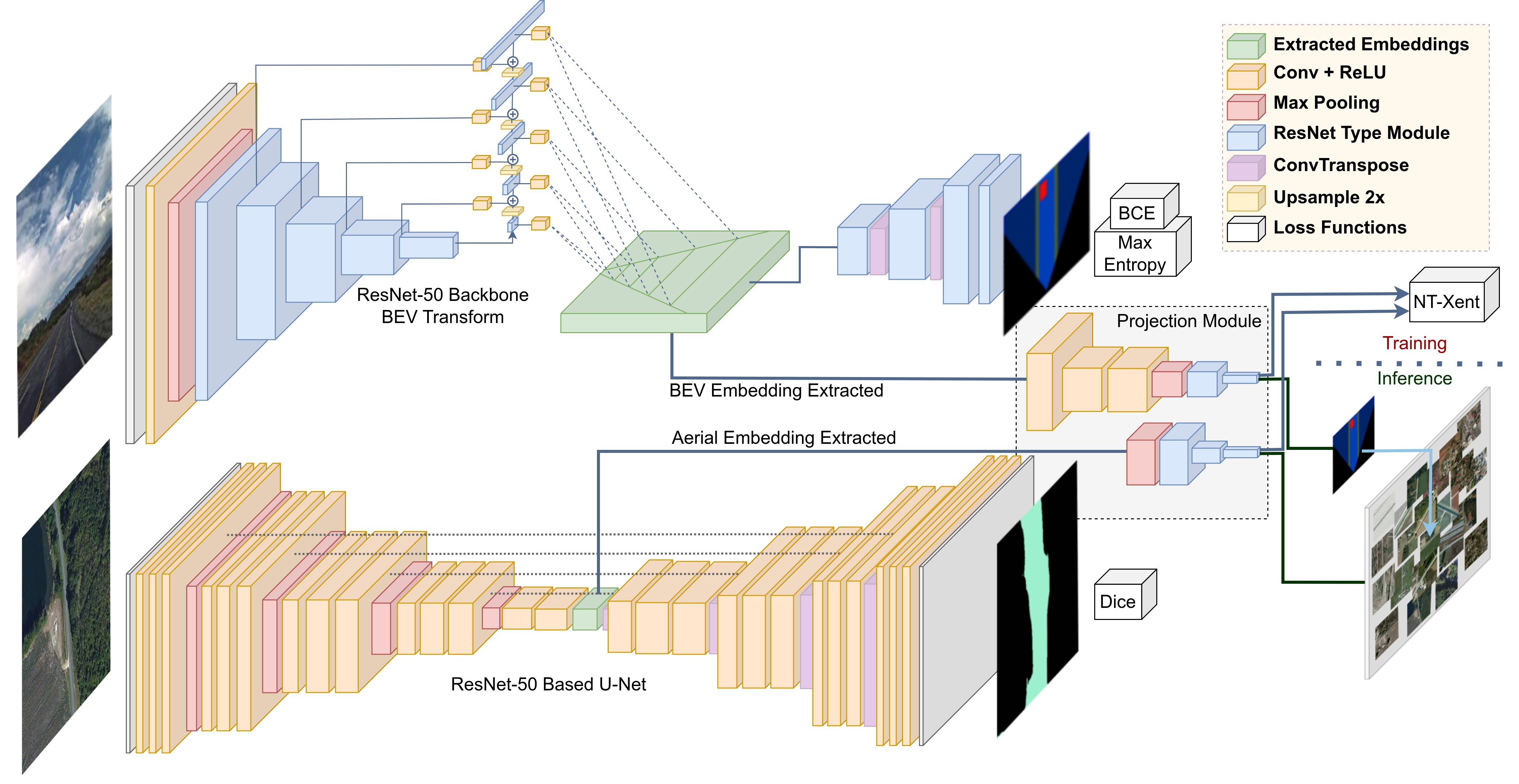}
    \caption{BEV-CV network overview: \textit{BEV Branch} is shown as the upper pathway, transforming from POV to BEV before extracting the embedding for projecting, the \textit{Aerial Branch} is the lower pathway, extracting embeddings from the U-Net latent space. At training time we use an NT-Xent loss function and at inference time we build a KDTree of aerial embedding and query this with POV embeddings using descriptor cosine similarity for retrieval.}
    \label{fig:arch}
\end{figure*}

\section{Methodology} 
The network's objective is to minimise the domain gap apparent between aerial and POV image viewpoints to produce similar embeddings from both inputs. We bring both images into the top-down view, extracting and projecting features into a shared representation space. The network architecture is shown in Figure \ref{fig:arch}, a two branch system with no weight sharing between branches.

\subsection{Semantic Feature Extraction}
To create a top-down view representation for the ground-level limited-FOV images we construct the \textit{BEV Branch} of BEV-CV. This network contains 4 stages to extract and re-sample perspective information across views. 

Both branches have an encoder-decoder structure to extract features that are reconstructed into semantic occupancy grids. Once trained on corresponding datasets, encoders are extracted and combined to form BEV-CV. We use semantic segmentation to compare embeddings as the classes layout within both images is similar, becoming comparable once transformed. %Previous SOTA methods \tav{}
Both POV and aerial input images are \\ RGB: $I_{t} \in \mathbb{R}^{3{\times}W{\times}H},  t \in \{pov, aer\}$.

\subsubsection{BEV Feature Extraction and Transform}
\begin{figure}[]
    \centering
    \includegraphics[width=\columnwidth]{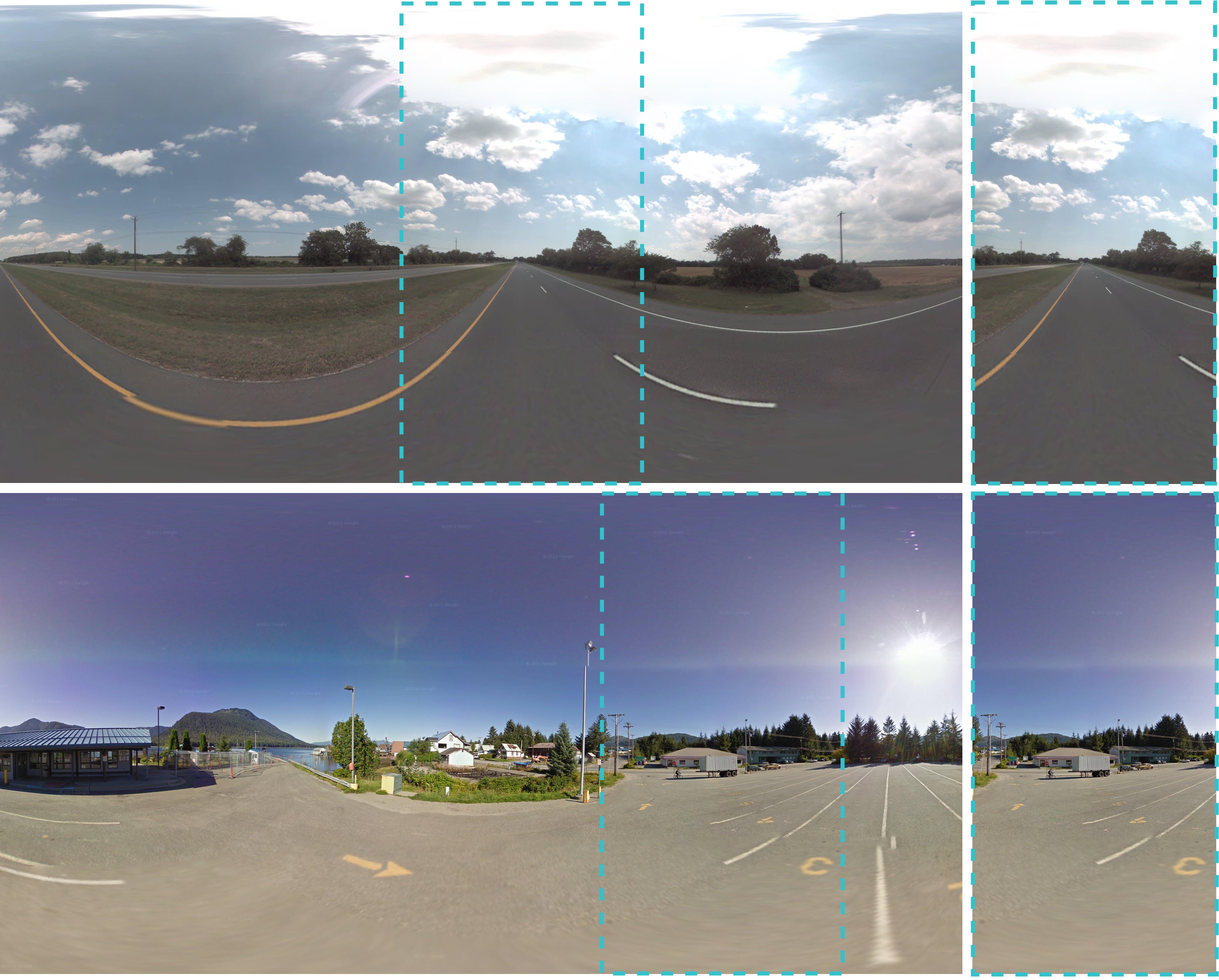}
    \caption{Panoramic examples of CVUSA and CVACT, heading aligned $90\degree$ FOV crops shown on the right hand side.}
    \label{fig:dataset}
\end{figure}

For the BEV, the input ground-level panoramic images are cropped to the limited-FOV (examples in Figure \ref{fig:dataset}) and resized to $224\times224$ where FOV, $\theta \in \{70\degree, 90\degree\}$, and yaw, $\Psi \in \{0\degree, ..., 360\degree\}$:
\begin{equation}
I_{pov} = \mathrm{fov\_crop}\left(I_{pov}, \theta, \Psi\right)\,.
\label{eq:crop}
\end{equation}

The images are fed through a ResNet-50 based network that extracts features at decreasing resolutions in order to retain information at different depths from the camera. These extractions are concatenated into a Feature Pyramid Network (FPN) which combines strong low-resolution semantics with weak high-resolution semantics through a top-down path and sequential lateral concatenations, shown as the \textit{BEV Transform} in Figure \ref{fig:arch}. $\mathrm{f_{i}}$ denotes the $i^{th}$ output for this $n$-layer FPN, concatenating ($\oplus$) outputs from the corresponding backbone layer with up-sampled lower-resolution features $f_{i-1}$: 
\begin{equation}
\mathrm{f_{i}}(I_{pov}) = \mathrm{conv}(R_{n-i}(I_{pov})) \oplus \mathrm{u}(f_{i-1}(I_{pov})) \label{eq:conv} 
\end{equation}
where $R_{n-i}$ are ResNet layer outputs at increasing depth, producing $n$ separate outputs as the feature pyramid. 

FPN activation values at pixel locations are re-sampled into the BEV mapping using a multi-scale dense transform (MSD) with camera intrinsics, $\rchi$, to expand features in the z-dimension with $\Delta{x}$ grid resolution, completing the conversion between the vertical to the horizontal plane. Re-sampling uses calibration intrinsics to determine the conversion of semantic information from object height to object depth. These transforms, $MSD_i$, condense features along the vertical dimension while maintaining horizontal resolution and expanding through depth, such that:
\begin{equation} 
\eta_{pov} = 
%\mathrm{BEV}(f_{i}, \chi) = 
\psi\left(\sum_{0}^{n}{\mathrm{MSD}_{i}\left(f_{i}, \frac{\chi_{f}\Delta x}{2^{n+3}}, \chi \right)}\right)
\label{eq:sample} 
\end{equation}
where $\rchi_{f}$ is the focal length of the camera. 
%This produces BEV feature maps 
%$M_{ortho} \in \mathbb{R}^{64{\times}100{\times}100}$ that finally pass through a top-down network based on ResNet but with ConvTranspose layers before building the semantic occupancy grid probabilities. 
In BEV-CV, the extracted orthographic features are passed through a sequence of convolutional layers $\psi$, to produce a compressed BEV embedding, $\eta_{pov} \in \mathbb{R}^{1{\times}512}$.

\subsubsection{Aerial Feature Extraction}
For the \textit{Aerial Branch} of the network, we construct a ResNet-50 based U-Net. We select this structure as U-Nets \cite{Ronneberger2015UNetCN} have become a predominant method for semantic segmentation, retaining information at decreasing resolutions to improve spatial context. The aerial branch is pre-trained with the full U-Net, using concatenation connections between the encoding and decoding stages. After pre-training, only the encoder is used within BEV-CV. Similar to the BEV branch, aerial embeddings are extracted with a set of progressively encoded maps, $\mathrm{e}_{0...n}$, with an equal number of decoders, $\mathrm{d}_{0...n}$, each deconvolving from the previous decoder and concatenating extractions from the corresponding encoder.
\begin{equation}
\mathrm{d_{i}}(I_{aer}) = e_{n-i}(I_{aer}) \oplus \mathrm{deconv}(\mathrm{d}_{i-1}(I_{aer})) \label{eq:unet} 
\end{equation}
where $\mathrm{deconv}$ represents $\mathrm{ConvTranspose2D}$.
Unlike the BEV transform, this network only outputs from the final convolution module, $\mathrm{d}_n$.

Detaching the encoders takes the outputs from the BEV after the multi-scale dense transforms and aerial outputs from the latent space of the U-Net. To complete the BEV feature extraction branch, this representation is compressed with further convolution layers to form practical embeddings - setting dimensionality with an optimal balance between discriminability and KDTree complexity, an important limitation when considering real-world deployment in mobile robotics. 
To transform both encoder outputs into a shared representation space with a standardised size we append a projection module. This contains fully connected layers, leaky ReLU functions, and batch normalisation. Aerial image embeddings $\left(\eta_{aer} \in \mathbb{R}^{1{\times}512}\right)$ are taken from the output of the last U-Net encoding layer $\mathrm{e}_n\left(I_{aer}\right)$.

% \cite{NIPS2016_6b180037}
\subsection{Normalised Temperature-scaled Cross Entropy Loss}
The triplet loss is used throughout CVGL research to bring positive image pairs closer together and push negative pairs further apart in the representation space. Training BEV-CV with a triplet loss function yielded satisfactory results. We utilise a normalised temperature-scaled cross-entropy loss (NT-Xent) function \cite{chen2020simple} for the problem instead. NT-Xent takes the same inputs as triplet loss: perspective images along with corresponding positive and negative aerial image pairs. A variety of techniques are used for determining negative pairs, often depending on the initial L2 distance between embeddings. Hard triplet mining uses embeddings closer to the anchor than the positive for negatives, for semi-hard triplet mining the negative is not closer to the anchor than the positive is, but it still has a positive loss. We don't explicitly select triplets for training, instead using every other aerial image within a batch as negative examples, leading to each batch of size $B$ having $B(B-1)$ negative examples.

The loss function for a batch of $n$ CVGL embedding pairs $\left(\eta_{pov}^{0...n}, \eta_{aer}^{0...n}\right)$ is:
\begin{equation} 
\mathcal{L}(\eta_{pov}^{i}, \eta_{aer}^{i}) = -log\frac{D(\eta_{pov}^{i}, \eta_{aer}^{i}) }{\displaystyle\sum^{n}_{k=1, k \neq i} D(\eta_{pov}^{i}, \eta_{aer}^{k}) }\,,
 \label{eq:ntx1} 
\end{equation}
where $D$ is the temperature ($\tau$) normalised cosine similarity.
\begin{equation} 
\mathcal{D}(i,j) = exp\left(\frac{i^Tj}{\tau||i||||j||} \right)\,
\label{eq:ntx2} 
\end{equation}

The final loss is computed across all positive pairs in a batch.

\section{Results}

\subsection{Datasets}\label{datasets}
The BEV transform and U-Net were pre-trained with individual datasets from each viewpoint. The BEV was trained with NuScenes \cite{nuscenes}, and the U-Net with \textit{Massachusetts Road Dataset} \cite{MnihThesis} - consisting of aerial images and road semantic segmentation masks. Finally fine-tuning the constructed BEV-CV for CVGL.

We evaluate with CVGL benchmark datasets Crossview USA (CVUSA) \cite{7410808} and CVACT \cite{8954224}. These datasets contain 35,532 POV-Aerial training pairs and 8,884 testing pairs. CVUSA aerial images have a resolution of 750x750 and ground-level panoramas of 1232x224. CVACT aerial images have a resolution of 1200x1200 and ground-level panoramas of 1664x832, all north-aligned. We use the evaluation protocol implemented by \cite{Shi2020WhereAI}, \cite{8954224}, and \cite{Shi2019OptimalFT}. CVUSA contains yaw at image capture time, allowing for front-facing limited-FOV crops - the expected input for a real-world application of CVGL for autonomous vehicles. Unlike previous works, we do not perform pre-processing such as aerial polar transforms or panoramic shifting crops to match orientation between the viewpoints. 
We use a more realistic evaluation protocol where images are aligned to the heading of the car - achievable simply using a cheap compass sensor and the vehicle's yaw. Thus we avoid the expensive pre-processing steps used by DSM \cite{Shi2020WhereAI} and GAL \cite{gal} to observe orientation-aware performance. We evaluate previous works with original publicly available source code, and our direct image alignment protocol.
CVACT does not contain the vehicle's yaw at image capture time. To evaluate with CVACT we first estimate the heading for each image pair using a semantic segmentation network \cite{zhou2018semantic} on the panoramic images. The yaw inaccuracies from this adversely affect results evaluated under the aligned protocol, however this would not be apparent in real-world application where vehicle-mounted cameras would be inherently aligned. 

\begin{table}[]
\centering
\begin{tabular}{l|ccccc}
Config (BEV-CV minus) & R@1 & R@5 & R@10 & R@1\% \\ \hline
- BEV & 0.80 & 3.11 & 5.53 & 21.35 \\
- Projection & 9.83 & 24.15 & 33.26 & 73.70 \\ 
- U-Net & 12.95 & 31.43 & 42.55 & 81.57 \\ \hline
Full BEV-CV & \textbf{14.03} & \textbf{32.32} & \textbf{43.25} & 81.48
\end{tabular}%
\caption{Ablation study with CVUSA 70$\degree$ crops.}
\label{tab:abla}
\end{table}

\begin{table}[]
\centering
\begin{tabular}{c|cccc}
 Offset & R@1 & R@5 & R@10 & R@1\% \\ \hline
0$\degree$ & \textbf{14.03} & \textbf{32.32} & \textbf{43.25} & \textbf{81.48} \\
$\pm$5$\degree$  & 11.89 & 26.74 & 36.37 & 74.41 \\
$\pm$15$\degree$ & 9.79 & 23.31 & 31.86 & 71.07 \\
$\pm$25$\degree$ & 8.81 & 20.14 & 28.13 & 66.19 \\
$\pm$35$\degree$ & 2.80 & 10.82 & 17.54 & 51.77 \\
$\pm$45$\degree$ & 1.11 & 5.36 & 9.70 & 39.18
\end{tabular}
\caption{Recall accuracies with offset 70$\degree$ POV crops}
\label{tab:offset}
\vspace{-2em}
\end{table}

\subsection{Training Details}
Training occurs in two stages: pre-training the BEV \& U-Net, and training the combined two-branch network. The BEV network is trained with NuScenes to take ground-level monocular images as input - outputting semantic map representations. The aerial branch's U-Net is trained with the Massachusetts dataset to extract road semantic features. The encoders from both networks are extracted, combining the branches to form a network trained with the evaluation datasets described in Section \ref{datasets}. BEV-CV is trained with triplets for 80 epochs using an Adam optimiser with an initial learning rate of 1e-4 and a \textit{ReduceLROnPlateau} scheduler. 

\subsection{Implementation Details}
We implement the architecture using PyTorch \cite{NEURIPS2019_bdbca288} with the Lightning framework \cite{Falcon_PyTorch_Lightning_2019}, and style our BEV network from \cite{Roddick2020PredictingSM}. Once trained, we extract the encoder consisting of a ResNet-50 backbone whose outputs at multiple resolutions are concatenated into a 5-layer Feature Pyramid Network (FPN) before entering multi-scale dense transforms as BEV features with shape $[64,100,100]$. We append 3 \textit{Conv-BatchNorm-LeakyReLU} sequences to compress feature extractions to shape $[512,7,7]$. We extract the U-Net encoder output, disregarding prior concatenation operations. The U-Net is a ResNet-50 based architecture with outputs features of shape $[2048,7,7]$. Feature extractions are projected into a shared representation space using a module that first applies a max pooling layer to flatten inputs. For the aerial branch, this is followed by a fully-connected layer to reduce the output size to 512. For both branches, the network ends with a single module of \textit{BatchNorm-LeakyReLU-FC} which leaves the dimensions unchanged but in the shared latent space.

\subsection{Evaluation}
We use Top-K recall accuracy to evaluate, similar to previous works \cite{Shi2020WhereAI}, \cite{8578856}, \cite{Shi2019OptimalFT}, \cite{gal}. Constructing a KDTree of aerial image embeddings, we retrieve the Top-K of these for a queried POV embedding determined by the cosine similarity between descriptors. A query is deemed successfully localised if the correct aerial image is within the top \textit{K} retrievals. Top-K uses the absolute value of K for retrievals whereas Top-K\% uses the K\% length of the total dataset.

\begin{figure}[]
    \centering
    \includegraphics[width=\columnwidth]{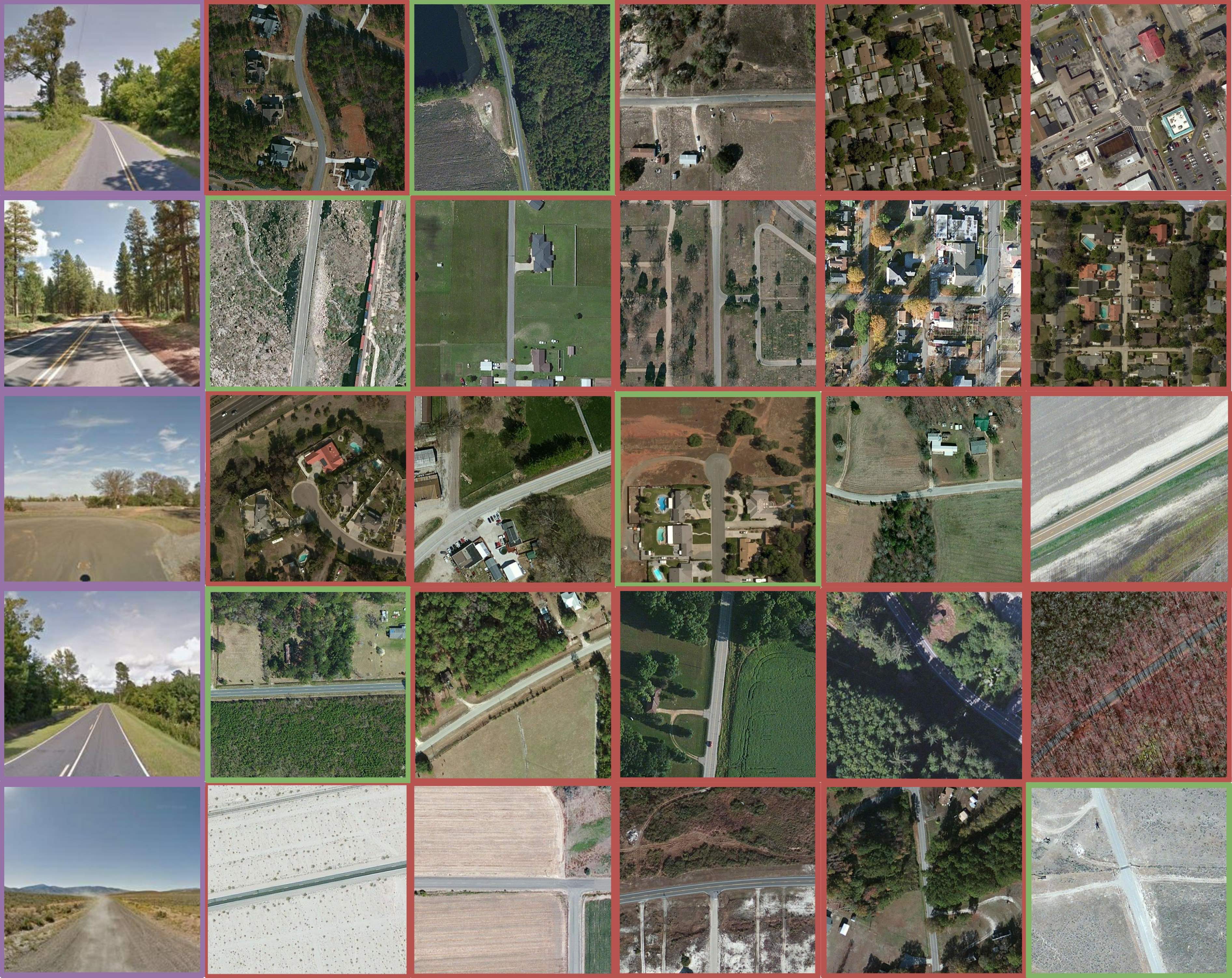}
    \caption{BEV-CV CVUSA Top-5 recall examples. Outlines: Purple - query POV image, green - correct aerial image, red - incorrect aerial image}
    \label{fig:qual}
\end{figure}

% Model sizes
% CVM-NET 1.8G
% CVFT 336.8M
% DSM 244.8M
%
%
%
% BEV-CV 259.8M

\subsection{Ablation Study}
To demonstrate the BEV module's effectiveness we ran an ablation study - replacing each branch with a vanilla ResNet-50, also removing the projection module, results are shown in Table \ref{tab:abla}. This concluded that each constituent module within BEV-CV has a positive impact on its performance, with the BEV transform causing the largest increase in Top-K accuracy. To display the sensitivity of BEV-CV to POV road-alignment, we add an offset angle to the yaw before cropping panoramas, the experimental outcome is shown in Table \ref{tab:offset}. Although BEV-CV performs well with misalignment, to operate optimally - POV images should be road-aligned.

% \begin{figure}[]
%     \centering
%     \includegraphics[width=0.7\columnwidth]{figures/tr_ntx.jpg}
%     \caption{Top-K accuracy comparisons between Triplet and NT-Xent losses.}
%     \label{fig:ntx}
% \end{figure}
\begin{table*}[]
\centering
\begin{tabular}{c|c|cccc|cccc}
\textbf{Model} & \textbf{\begin{tabular}[c]{@{}c@{}}Orientation\\ Aware\end{tabular}} & \textbf{R@1} & \textbf{R@5} & \textbf{R@10} & \textbf{R@1\%} & \textbf{R@1} & \textbf{R@5} & \textbf{R@10} & \textbf{R@1\%} \\ \hline
 &  & \multicolumn{4}{c|}{CVUSA $90\degree$} & \multicolumn{4}{c}{CVUSA $70\degree$} \\
CVM \cite{8578856} & \xmark & 2.76 & 10.11 & 16.74 & 55.49 & 2.62 & 9.30 & 15.06 & 21.77 \\
CVFT \cite{Shi2019OptimalFT} & \xmark & 4.80 & 14.84 & 23.18 & 61.23 & 3.79 & 12.44 & 19.33 & 55.56 \\
DSM \cite{Shi2020WhereAI} & \xmark & 16.19 & 31.44 & 39.85 & 71.13 & 8.78 & 19.90 & 27.30 & 61.20 \\
L2LTR \cite{Yang2021CrossviewGW} & \xmark & 26.92 & 50.49 & 60.41 & 86.88 & 13.95 & 33.07 & 43.86 & 77.65\\
TransGeo \cite{Zhu2022TransGeoTI} & \xmark & 30.12 & 54.18 & 63.96 & 89.18 & 16.43 & 37.28 & 48.02 & 80.75 \\
GeoDTR \cite{zhang2023crossview} & \xmark & 18.81 & 43.36 & 57.94 & 88.14 & 14.84 & 38.03 & 51.27 & 88.17 \\
BEV-CV & \xmark & 15.17 & 33.91 & 45.33 & 82.53 & 14.03 & 32.32 & 43.25 & 81.48 \\ \hline
GAL \cite{gal} & $\bm{\approx}$ & 22.54 & 44.36 & 54.17 & 84.59 & 15.20 & 32.86 & 42.06 & 75.21 \\
DSM \cite{Shi2020WhereAI} & \cmark & \textbf{33.66} & 51.70 & 59.68 & 82.46 & 20.88 & 36.99 & 44.70 & 71.10 \\
L2LTR \cite{Yang2021CrossviewGW} & \cmark & 25.21 & \textit{51.90} & \textit{63.54} & \textit{91.16} & \textit{22.20} & \textit{46.71} & \textit{58.99} & \textit{89.37} \\
TransGeo \cite{Zhu2022TransGeoTI} & \cmark & 21.96 & 45.35 & 56.49 & 86.80 & 17.27 & 38.95 & 49.44 & 81.34 \\
GeoDTR \cite{zhang2023crossview} & \cmark & 15.21 & 39.32 & 52.27 & 88.72 & 14.00 & 35.28 & 47.77 & 86.39\\
% CVUSA 70 aligned GeoDTR 25.61, top_5: 53.25, top_10: 65.29, top_1_percent: 93.01
% 
BEV-CV & \cmark & \textit{32.11} & \textbf{58.36} & \textbf{69.06} & \textbf{92.99} & \textbf{27.40} & \textbf{52.94} & \textbf{64.47} & \textbf{90.94} \\ \hhline{==========}
 &  & \multicolumn{4}{c|}{CVACT $90\degree$} & \multicolumn{4}{c}{CVACT $70\degree$} \\
CVM \cite{8578856} & \xmark & 1.47 & 5.70 & 9.64 & 38.05 & 1.24 & 4.98 & 8.42 & 34.74 \\
CVFT \cite{Shi2019OptimalFT} & \xmark & 1.85 & 6.28 & 10.54 & 39.25 & 1.49 & 5.13 & 8.19 & 34.59 \\
DSM \cite{Shi2020WhereAI} & \xmark & 18.11 & 33.34 & 40.94 & 68.65 & 8.29 & 20.72 & 27.13 & 57.08 \\
L2LTR \cite{Yang2021CrossviewGW} & \xmark & 13.07 & 30.38 & 41.00 & 76.07 & 6.67 & 15.94 & 23.45 & 49.37 \\
TransGeo \cite{Zhu2022TransGeoTI} & \xmark & 10.75 & 28.22 & 37.51 & 70.15 & 7.01 & 19.44 &27.50 & 62.19  \\ 
GeoDTR \cite{zhang2023crossview} & \xmark & 26.53 & 53.26 & 64.59 & 91.13 & 16.87 & 40.22 & 53.13 & 87.92 \\
BEV-CV & \xmark & 4.14 & 14.46 & 22.64 & 61.18 & 3.92 & 13.50 & 20.53 & 59.34 \\ \hline
GAL \cite{gal} & $\bm{\approx}$ & 26.05 & 49.23 & 59.26 & \textit{85.60} & 14.17 & 32.96 & 43.24 & 77.49 \\
DSM \cite{Shi2020WhereAI} & \cmark & 31.17 & 51.44 & 60.05 & 82.90 & 18.44 & 35.87 & 44.39 & 71.97 \\
L2LTR \cite{Yang2021CrossviewGW} & \cmark & \textit{33.62} & 46.28 & 58.21 & 78.62 & \textit{28.65} & \textit{53.59} & \textit{65.02} & \textit{90.48} \\
TransGeo \cite{Zhu2022TransGeoTI} & \cmark & 28.16 & 34.44 & 41.54 & 67.15 & 24.05 & 42.68 & 55.47 & 80.72  \\
GeoDTR \cite{zhang2023crossview} & \cmark & 26.76 & \textit{53.65} & \textit{65.35} & \textit{92.12} & 15.38 & 37.09 & 49.40 & 86.38 \\
% GeoDTR CVACT 90 aligned 47.50 & 74.89 & 82.81 & 95.85 & 33.53 & 62.63 & 73.67 & 93.88 \\
BEV-CV & \cmark & \textbf{45.79} & \textbf{75.85} & \textbf{83.97} & \textbf{96.76} & \textbf{37.85} & \textbf{69.00} & \textbf{78.52} & \textbf{95.03} \\ \hline
\end{tabular}%
\caption{BEV-CV evaluation against previous works - focusing on road-aligned orientation-aware analysis, best results are shown in \textbf{bold}, with second best in \textit{Italic}. $\bm{\approx}$ denotes where code was unavailable and orientation was partly utilised.}
\label{tab:sota}
\vspace{-0.5em}
\end{table*}

\subsection{Triplet vs NT-Xent Loss}
Triplet loss has been widely used within computer vision research, the addition of a temperature parameter ($\tau$) in NT-Xent to scale cosine similarities has been found to improve learning from hard negative examples \cite{chen2020simple}. 
Utilising the NT-Xent loss function yielded improvements across all Top-K recall accuracies, on average by 5\%. With CVUSA $70\degree$ unaligned images, Top-1 and Top-1\% increased from 11.30\% to 14.03\% and from 74.4\% to 81.48\%, respectively.

\begin{table}[]
\centering
\resizebox{\columnwidth}{!}{
\begin{tabular}{c|cc|cc}
Model & Backbone & Dims $\downarrow$ & Params (M) $\downarrow$ & FLOPs (G) $\downarrow$ \\ \hline
CVM \cite{8578856} & VGG16 & 4096 & 160.3 & - \\
CVFT \cite{Shi2019OptimalFT} & VGG16 & 4096 & 26.8 & - \\
DSM \cite{Shi2020WhereAI} & VGG16 & 4096 & \textbf{14.5} & 39.3 \\
GAL \cite{gal} & ResNet-18 & - & - & - \\
L2LTR \cite{Yang2021CrossviewGW} & HybridViT & 768 & 195.9 & 57.1 \\
TransGeo \cite{Zhu2022TransGeoTI} & DeiT-S/16 & 1000 & 44.9 & 12.3 \\ 
GeoDTR \cite{zhang2023crossview} & ResNet-34 & 4096 & 48.5 & 39.89 \\ \hline
BEV-CV & ResNet-50 & \textbf{512} & 65.2 & \textbf{11.5}
\end{tabular}
}
\caption{Complexity comparison for real-time application feasibility.} 
\label{tab:params}
\vspace{-2.5em}
\end{table}

\subsection{Prior Work Comparison}
Recall rates are compared to previous SOTA techniques shown in Table \ref{tab:sota}. We focus on orientation-aware evaluation, where images are yaw aligned, as it is the most applicable design for real-world deployment - including the orientation-unaware evaluations for comparison and to demonstrate how previous works are effected by the change in data representation.
We achieve a 23\% improvement in Top-1 rates with 70$\degree$ CVUSA crops, and 24\% improvement in Top-1 rates with 70$\degree$ CVACT crops, proving the utility of BEV transforms for reducing the \ac{cvgl} domain disparity. 
\vspace{-0.5em}

\subsection{Computational Efficiency}
Comparing against these works, we also demonstrate computation improvements, shown in Table \ref{tab:params}. Reducing dimensionality dramatically improves real-world feasibility as querying for Top-K retrievals in a KDTree takes at best $O(\sqrt{D} + k)$ time, where $k$ is the number of retrievals, and $D$ the dimensionality of the KDTree. Therefore lowering the dimensionality from 768-dims to 512-dims reduces Top-1 query retrieval time complexity and KDTree memory requirements by 18\% and 33\% respectively. We reduce \ac{flops} by 6.5\% compared with the next best previous work - further increasing feasibility for use in mobile robotics where computational capacity is limited and expensive.  

\section{Conclusion \&  Future Work}
\vspace{-0.5em}
This paper introduced a novel technique to reduce the domain gap within limited-FOV CVGL, establishing the validity of BEV transforms in CVGL as a route to increase real-world feasibility. 
Evaluating against previous feature extraction approaches with road-aligned image queries, we improve upon nearly all recall accuracies for both CVUSA and CVACT, demonstrating strong potential for the field. BEV-CV achieves this while reducing the computational requirements for practical implementation - lowering both image embedding dimensionalities and retrieval time complexities. 
Our approach has some limitations; for instance, BEV transform specifications are set during training, determining parameters and transform shape within the BEV transform module according to the camera intrinsics. This limits the generalisability of the network for unseen image intrinsics at inference. Recent advances in BEV transforms have attempted to remove the need for explicit intrinsics, this will be a valuable addition to future BEV-CV techniques.
However, we have shown that CNN-based BEV networks can be used as a drop-in replacement for the backbones commonly used in CVGL. 
Further work should aim to generalise BEV-CV across a wider varieties of regions, light, and weather conditions as current datasets were collected during daytime with clear weather from a small region.

\section{Acknowledgements}
This work was partially funded by the EPSRC under grant agreement EP/S035761/1 and FlexBot - InnovateUK project 10067785.

\newpage
\addtolength{\textheight}{-17cm}   % This command serves to balance the column lengths

%%%%%%%%%%%%%%%%%%%%%%%%%%%%%%%%%%%%%%%%%%%%%%%%%%%%%%%%%%%%%%%%%%%%%%

%%%%%%%%%%%%%%%%%%%%%%%%%%%%%%%%%%%%%%%%%%%%%%%%%%%%%%%%%%%%%%%%%%%%%%%%%%%%%%%%

\printbibliography

\end{document}